\documentclass[10pt,twocolumn,letterpaper]{article}

\usepackage{cvpr}
\usepackage{times}
\usepackage{epsfig}
\usepackage{graphicx}
\usepackage{amsmath}
\usepackage{amssymb}
\usepackage{comment}
\usepackage{subfigure}

\usepackage{booktabs}
\usepackage{array, makecell, tabularx}
\usepackage{xspace}

\usepackage{subfigure}
\usepackage{xcolor}
\usepackage{rotating}
\usepackage[export]{adjustbox}

\usepackage[pagebackref=true,breaklinks=true,letterpaper=true,colorlinks,bookmarks=false]{hyperref}

\newcommand{\approxtilde}{{\raise.17ex\hbox{$\scriptstyle\sim$}}}

\makeatletter
\def\@rothead[#1]#2{\thead{\\[-.65\normalbaselineskip]
  \turn{\cellrotangle}\thead[#1]{#2}\endturn}}
\makeatother

\cvprfinalcopy %

\def\cvprPaperID{05} %

\ifcvprfinal\fi

\newcommand*\samethanks[1][\value{footnote}]{\footnotemark[#1]}

\begin{document}

\title{Training Deep Networks with Synthetic Data: \\Bridging the Reality Gap by Domain Randomization}

\author{
    Jonathan Tremblay\thanks{equal contribution}\\
    \and
    Aayush Prakash\samethanks\\
    \and
    David Acuna\samethanks$\hspace{0.7ex}^\dagger$ \\
    \and
    Mark Brophy\samethanks\\
    \and
    Varun Jampani\\
    \and
    Cem Anil$^\dagger$ \\
    \and
    Thang To\\
    \and
    Eric Cameracci\\
    \and
    Shaad Boochoon\\
    \and
    Stan Birchfield\\
		\and 
\hspace{10ex} \parbox[c][2.9ex]{45ex}{\centering NVIDIA} \\
		\and \and \and \and
\parbox{40ex}{\centering $^\dagger$also University of Toronto} \\
		\and
    {\tt\small \{jtremblay,aayushp,dacunamarrer,markb,vjampani,} \\
		\hspace{5ex} {\tt\small thangt,ecameracci,sboochoon,sbirchfield\}@nvidia.com} \\
    {\tt\small cem.anil@mail.utoronto.com}
}

\maketitle
\begin{abstract}

We present a system for training deep neural networks for object detection using synthetic images.
To handle the variability in real-world data,
the system relies upon the technique of domain randomization,
in which the parameters of the simulator---such as lighting, pose, object textures, {\em etc.}---are randomized in non-realistic ways to force the neural network to learn the essential features of the object of interest.
We explore the importance of these parameters,
showing that it is possible to produce a network with compelling performance using only non-artistically-generated synthetic data.
With additional fine-tuning on real data,
the network yields better performance than using real data alone.
This result opens up the possibility of using inexpensive synthetic data
for training neural networks while avoiding the need to collect
large amounts of hand-annotated real-world data
or to generate high-fidelity synthetic worlds---both of which remain bottlenecks for many applications.
The approach is evaluated on bounding box detection of cars on the KITTI 
dataset.
\end{abstract}
\section{Introduction}
\label{sec:introduction}

Training and testing a deep neural network is a time-consuming and expensive task which typically involves collecting and manually annotating a large amount of data for supervised learning.
This requirement is problematic when the task demands either expert knowledge, labels that are difficult to specify manually, or images that are difficult to capture in large quantities with sufficient variety.
For example, 3D poses or pixelwise segmentation can take 
a substantial amount of time for a human to manually label a single image.

A promising approach to overcome this limitation
is to use a graphic simulator to generate automatically labeled data.
Several such simulated datasets have been created in recent years
\cite{butler2012eccv,handa2015arx:sn,DFIB15iccv,mayer2015arx,qiu2016arx:uncv,zhang2016arx:unst, mccormac2016arx:snrgbd,ros2016cvpr:syn,Richter_2016_ECCV,gaidon2016CVPR,Mueller2017ue4,Tsirikoglou2017exporter}.
For the most part, these datasets are expensive to generate, requiring
artists to carefully model specific environments in detail.
These datasets have been used successfully for training networks for geometric problems such as optical flow,
scene flow, stereo disparity estimation, and camera pose estimation.

Even with this unprecedented access to high-fidelity ground truth data,
it is not obvious how to effectively use such data to train neural networks to operate on real images.  In particular, the expense required to generate photorealistic quality negates the primary selling point of synthetic data, namely, that arbitrarily large amounts of labeled data are available essentially for free.
To solve this problem,
\emph{domain randomization} \cite{tobin17:dr} is a recently proposed
inexpensive approach that intentionally abandons photorealism
by randomly perturbing the environment in non-photorealistic ways ({\em e.g.}, by adding random textures) to force the network to learn to focus on the essential features of the image.
This approach has been shown successful in tasks such as
detecting the 3D coordinates of homogeneously colored cubes on a table \cite{tobin17:dr} or determining the control commands of an indoor quadcopter~\cite{SadeghiL2017rss}, as well as for optical flow~\cite{DFIB15iccv} and scene flow~\cite{mayer2015arx}.

In this paper we extend domain randomization (DR)
to the task of detection of real-world objects.
In particular, we are interested in answering the following questions:
1) Can DR on synthetic data achieve compelling results on real-world data?
2) How much does augmenting DR with real data during training improve accuracy?
3) How do the parameters of DR affect results?
4) How does DR compare to higher quality/more expensive synthetic datasets?
In answering these questions, this work contributes the following:
\begin{itemize}
\itemsep-0.3em
\item Extension of DR to non-trivial tasks
such as detection of real objects in front of complex backgrounds;
\item Introduction of a new DR component, namely, \emph{flying distractors}, which
improves detection / estimation accuracy;
\item Investigation of the parameters of DR to evaluate their importance for these tasks;
\end{itemize}

As shown in the experimental results, we achieve competitive results on real-world tasks when trained using only synthetic DR data.  For example, our DR-based car detector achieves better results on the KITTI dataset than the same architecture trained on virtual KITTI \cite{gaidon2016CVPR}, even though the latter dataset is highly correlated with the test set.  Furthermore, augmenting synthetic DR data by fine-tuning on real data yields better results than training on real KITTI data alone.

\section{Previous Work}
\label{sec:previous}

The use of synthetic data for training and testing deep neural networks has gained in popularity in recent years,
as evidenced by the availability of a large number of such datasets:
Flying Chairs~\cite{DFIB15iccv},
FlyingThings3D~\cite{mayer2015arx},
MPI Sintel~\cite{butler2012eccv},
UnrealStereo \cite{qiu2016arx:uncv,zhang2016arx:unst},
SceneNet \cite{handa2015arx:sn},
SceneNet RGB-D \cite{mccormac2016arx:snrgbd},
SYNTHIA \cite{ros2016cvpr:syn},
GTA V \cite{Richter_2016_ECCV},
Sim4CV \cite{Mueller2017ue4},
and Virtual KITTI~\cite{gaidon2016CVPR}, among others.
These datasets were generated for the purpose of geometric problems
such as optical flow, scene flow, stereo disparity estimation,
and camera pose estimation.

Although some of these datasets also contains labels for object detection and semantic segmentation,
few networks trained only on synthetic data for these tasks have appeared.
Hinterstoisser {\em et al.}
\cite{hinterstoisser2017arx:pretrain} used synthetic data generated by adding Gaussian noise to the object of interest and Gaussian blurring the object edges before composing over a background image.
The resulting synthetic data are used to train the later layers of a neural network while freezing the early layers pretrained on real data (\emph{e.g.}, ImageNet).
In contrast, we found this approach of freezing the weights to be harmful rather than helpful, as shown later.

The work of Johnson-Roberson \etal \cite{johnson2017icra:ditmatrix} used photorealistic synthetic data to train a car detector that was tested on the KITTI dataset.  This work is closely related to ours, with the primary difference being our use of domain randomization rather than photorealistic images.  Our experimental results reveal a similar conclusion, namely, that synthetic data can rival, and in some cases beat, real data for training neural networks.  Moreover, we show clear benefit from fine-tuning on real data after training on synthetic data. 
Other non-deep learning work that uses intensity edges from synthetic 3D models to detect isolated cars can be found in \cite{stark2010bmvc:bttf}.

As an alternative to high-fidelity synthetic images, domain randomization (DR) was introduced by Tobin {\em et al.} \cite{tobin17:dr},
who propose to close the reality gap by generating synthetic data with sufficient variation that the network views real-world data as just another variation.
Using DR, they trained a neural network to estimate the 3D world position of various
shape-based objects with respect to a robotic arm fixed to a table.
This introduction of DR was inspired by the earlier work of Sadeghi and Levine
\cite{SadeghiL2017rss}, who trained a quadcopter to fly indoors using only synthetic images.  
The Flying Chairs \cite{DFIB15iccv} and FlyingThings3D \cite{mayer2015arx} datasets for optical flow and scene flow algorithms can be seen as versions of domain randomization.

The use of DR has also been explored to train robotics control policies.  The network of James {\em et al.} \cite{Jamesdr2017} 
was used to cause a robot to pick a red cube and place it in a basket, and 
the network of Zhang {\em et al.} \cite{Zhangdr2017} was used to  
position a robot near
a blue cube.
Other work explores
learning robotic policies from a high-fidelity rendering engine~\cite{james_nips_2016},
generating high-fidelity synthetic data via a procedural approach~\cite{Tsirikoglou2017exporter},
or training object classifiers from 3D CAD models~\cite{peng2015iccv:ldod}.
In contrast to this body of research, our goal is to use synthetic data to train networks that detect complex, real-world objects.

A similar approach to domain randomization is to paste real images (rather than synthetic images) of objects on background images, as proposed by Dwibedi {\em et al.} \cite{dwibedi2017iccv:cutpaste}.  One challenge with this approach is the accurate segmentation of objects from the background in a time-efficient manner.
\section{Domain Randomization}
\label{sec:method}

\begin{figure*}
    \includegraphics[width=2\columnwidth]{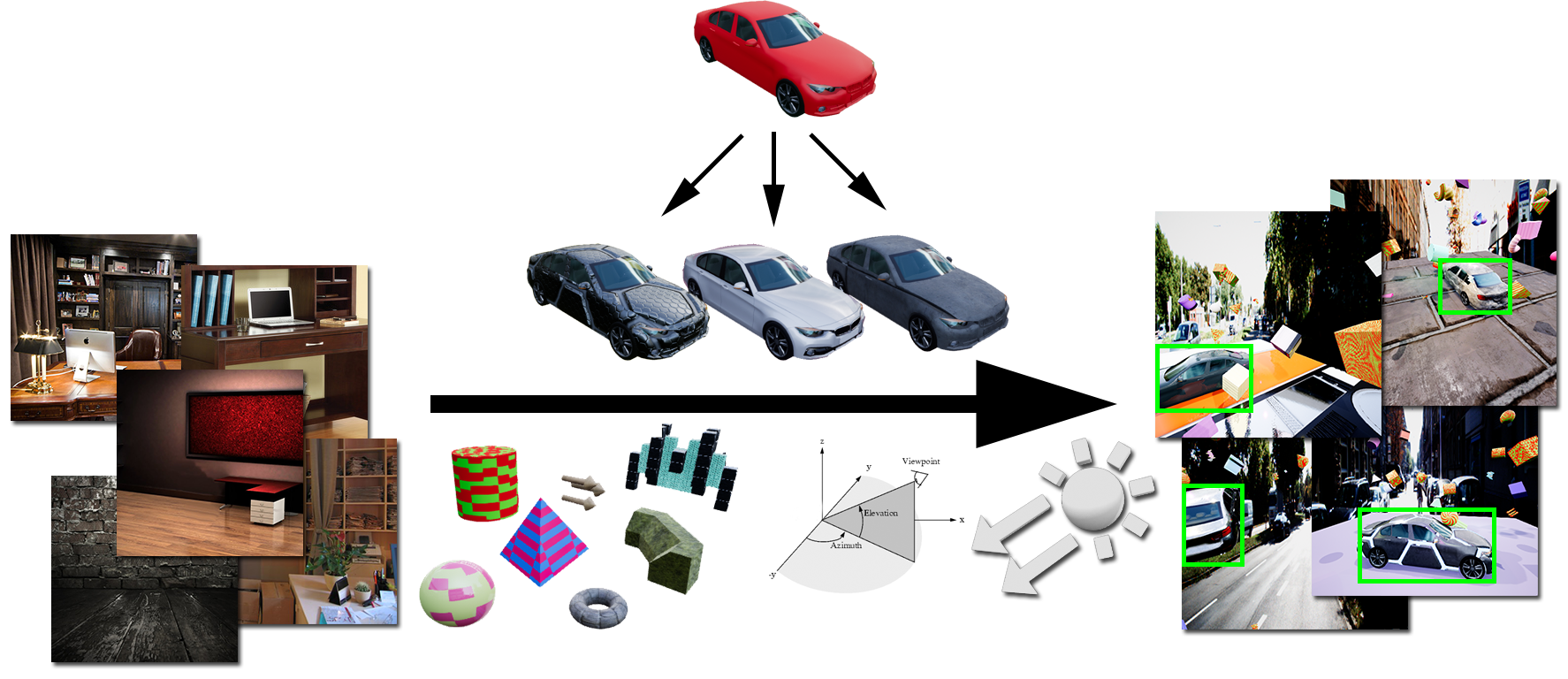}
    \caption{Domain randomization for object detection.
		Synthetic objects (in this case cars, top-center) are rendered on top of a random background (left) along with random flying distractors (geometric shapes next to the background images) in a scene with random lighting from random viewpoints.
		Before rendering, random texture is applied to the objects of interest as well as to the flying distractors.
		The resulting images, along with automatically-generated ground truth (right), are used for training a deep neural network.
		}
    \label{fig:dr_exporter}
\end{figure*}

Our approach to using domain randomization (DR) to generate synthetic data for training a neural network is illustrated in Fig.~\ref{fig:dr_exporter}.  We begin with 3D models of objects of interest (such as cars).  A random number of these objects are placed in a 3D scene at random positions and orientations.  To better enable the network to learn to ignore objects in the scene that are not of interest, a random number of geometric shapes are added to the scene.  We call these \emph{flying distractors}.  Random textures are then applied to both the objects of interest and the flying distractors.  A random number of lights of different types are inserted at random locations, and the scene is rendered from a random camera viewpoint, after which the result is composed over a random background image.  The resulting images, with automatically generated ground truth labels (e.g., bounding boxes), are used for training the neural network.

More specifically, images were generated by randomly varying the following aspects of the scene:
\vspace{-0.8em}
\begin{itemize}
\setlength\itemsep{-0.3em}
  \item number and types of objects, selected from a set of 36 downloaded 3D models of generic sedan and hatchback cars;  
	\item number, types, colors, and scales of distractors, selected from a set of 3D models (cones, pyramids, spheres, cylinders, partial toroids, arrows, pedestrians, trees, etc.);
	\item texture on the object of interest, and background photograph, both taken from the Flickr~8K~\cite{Hodosh2013} dataset;
	\item location of the virtual camera with respect to the scene (azimuth from $0^\circ$ to $360^\circ$, elevation from $5^\circ$ to $30^\circ$);
	\item angle of the camera with respect to the scene (pan, tilt, and roll from $-30^\circ$ to $30^\circ$);
	\item number and locations of point lights (from 1 to 12), in addition to a planar light for ambient illumination;
	\item visibility of the ground plane.
	\end{itemize}

Note that all of these variations, while empowering the network to achieve more complex behavior, are nevertheless extremely easy to implement, requiring very little additional work beyond previous approaches to DR.
Our pipeline uses an internally created plug-in to the Unreal Engine (UE4) that is capable of outputing $1200 \times 400$ images with annotations at 30~Hz.

\begin{figure*}
    \centering
    \begin{tabular}{ccc}
      \includegraphics[width=0.3\textwidth]{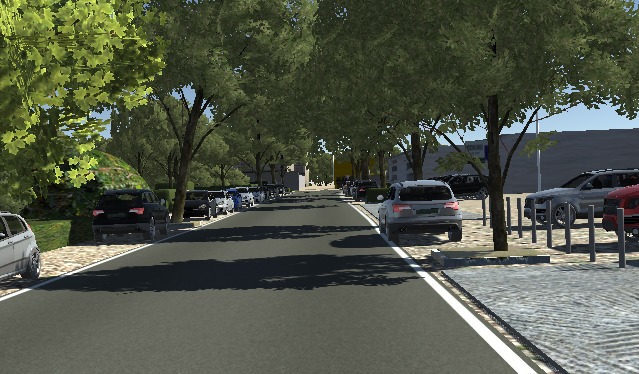} &
      \includegraphics[width=0.3\textwidth]{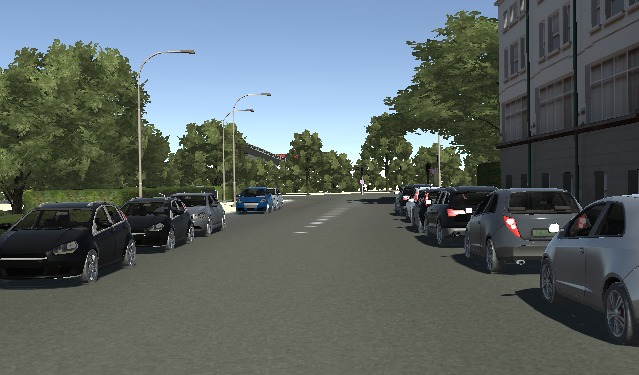} &
      \includegraphics[width=0.3\textwidth]{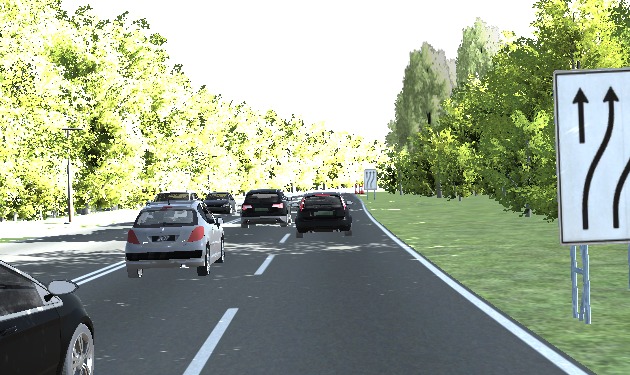} \\

      \includegraphics[width=0.3\textwidth]{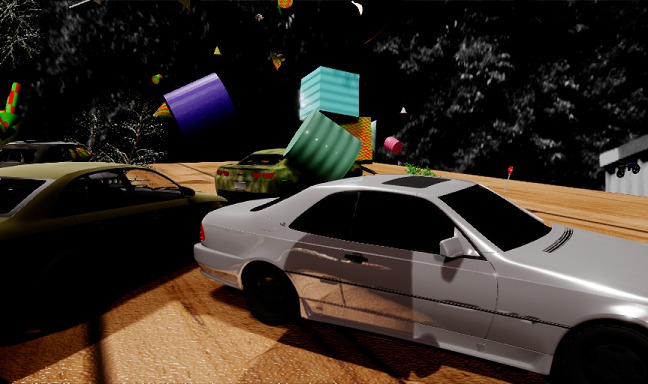} &
      \includegraphics[width=0.3\textwidth]{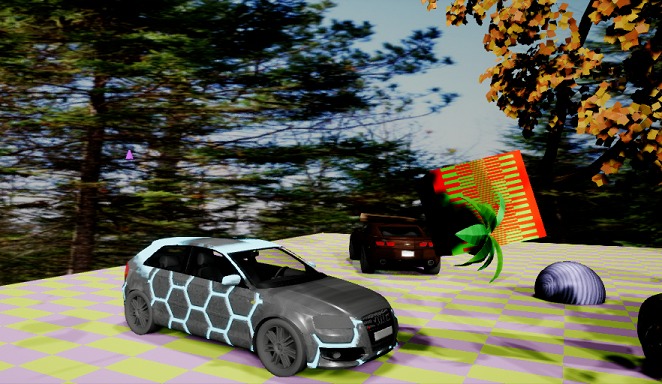} &

      \includegraphics[width=0.3\textwidth]{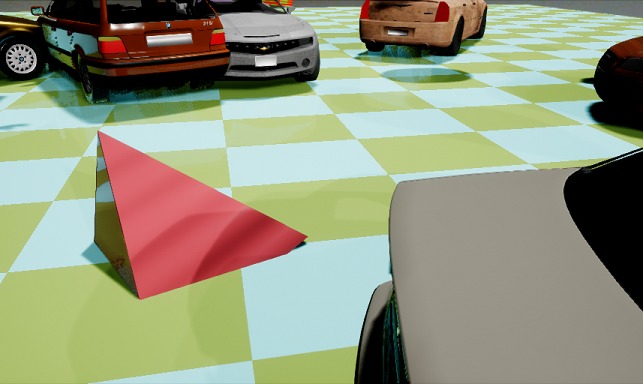} \\

    \end{tabular}
    \caption{Sample images from Virtual KITTI (first row), and our DR approach (second row).  Note that our images are easier to create (because of their lack of fidelity) and yet contain more variety to force the deep neural network to focus on the structure of the objects of interest.}
    \label{fig:training}
\end{figure*}

A comparison of the synthetic images generated by our version of DR with the high-fidelity Virtual KITTI (VKITTI) dataset \cite{gaidon2016CVPR} is shown in 
Fig.~\ref{fig:training}.  
Although our crude (and almost cartoonish) images are not as aesthetically pleasing, this apparent limitation is arguably an asset:  Not only are our images orders of magnitude faster to create (with less expertise required) but they include variations that force the deep neural network to focus on the important structure of the problem at hand rather than details that may or may not be present in real images at test time.

\section{Evaluation}\label{sec:results}

\begin{figure}
    \hspace*{-5mm}
    \centering
    \begin{tabular}{cc}

      \includegraphics[width=0.49\columnwidth]{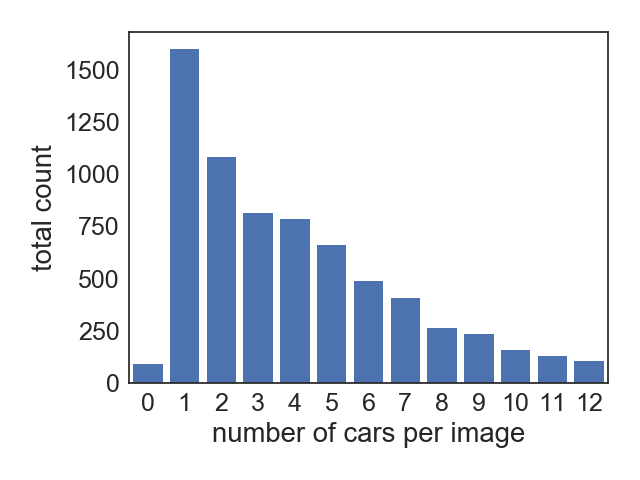} &
      \includegraphics[width=0.49\columnwidth]{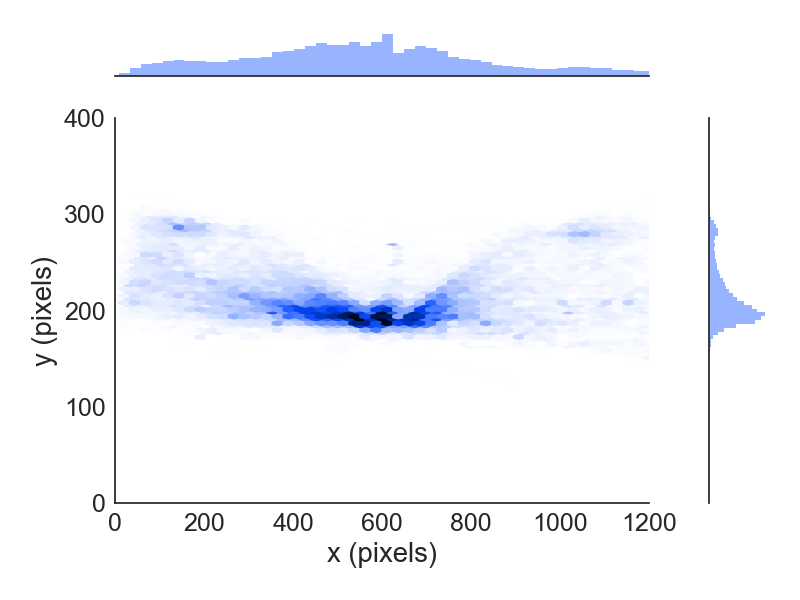} \\
      \includegraphics[width=0.49\columnwidth]{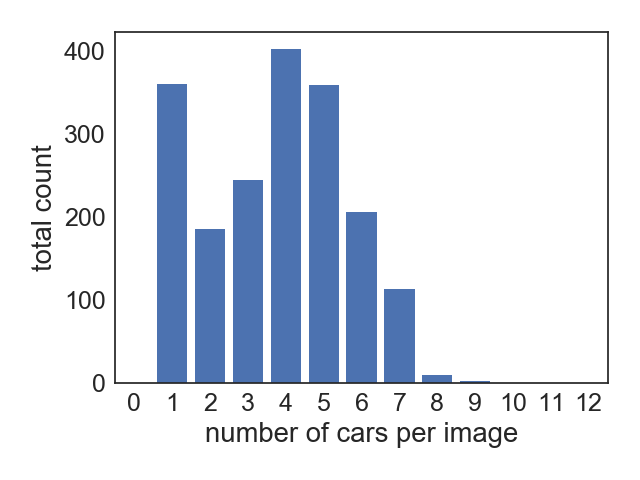} &
      \includegraphics[width=0.49\columnwidth]{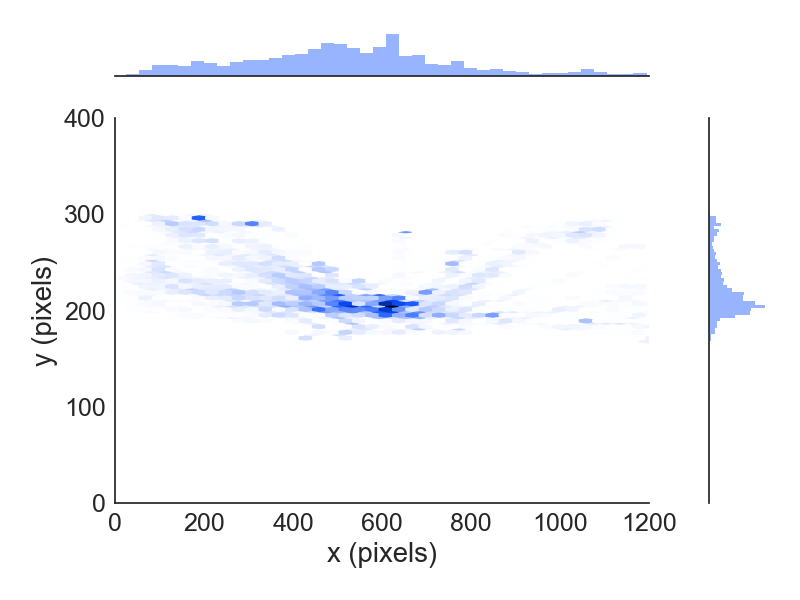} \\
      \includegraphics[width=0.49\columnwidth]{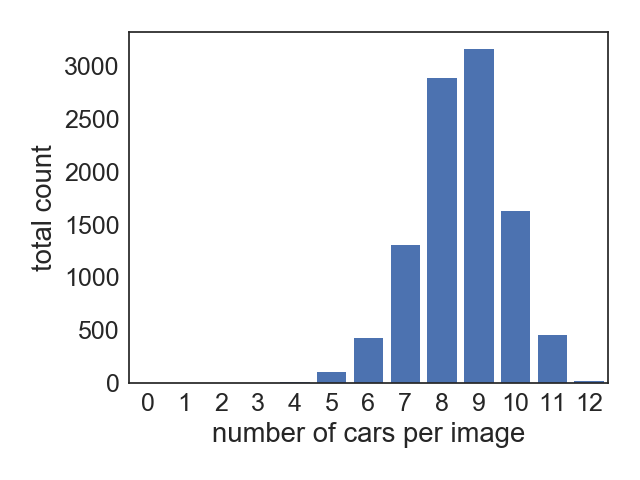} &
      \includegraphics[width=0.49\columnwidth]{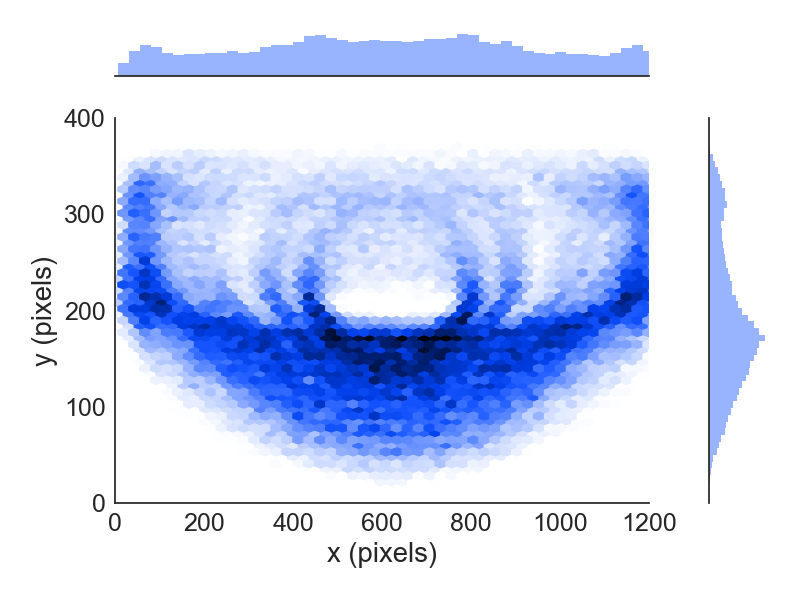} \\

    \end{tabular}
    \caption{Statistics of the KITTI (top), virtual KITTI (middle), and our DR dataset (bottom).  
		Shown are the distributions of the number of cars  
    per image (left) and car centroid location (right).
		Note that with DR it is much easier to generate data with a wide distribution.
    }
    \label{fig:stats_dataset}
\end{figure}

To quantify the performance of domain randomization (DR), in this section we compare the results of training an object detection deep neural network (DNN) using images generated by our DR-based approach with results of the same network trained using synthetic images from
the Virtual KITTI (VKITTI) dataset \cite{gaidon2016CVPR}.
The real-world KITTI dataset \cite{Geiger2012CVPR} was used for testing.
Statistical distributions of these three datasets are shown in Fig.~\ref{fig:stats_dataset}. 
Note that our DR-based approach makes it much easier to generate a dataset with a large amount of variety,
compared with existing approaches.

\subsection{Object detection}
\label{sec:detection}

We trained three state-of-the-art neural networks using open-source implementations.\footnote{\url{https://github.com/tensorflow/models/tree/master/research/slim}}
In each case we used the feature extractor recommended by the respective authors.  These three network architectures are briefly described as follows.

\vspace{2mm}\noindent\textbf{Faster R-CNN} \cite{faster-rcnn2015NIPS} detects objects in two stages.  The first stage is a region proposal network (RPN) that generates candidate regions of interest using extracted features along with the likelihood of finding an object in each of the proposed regions.
In the second stage, features are cropped from the image using the proposed regions and fed to the remainder of the feature extractor, which predicts a probability density function over object class along with a refined class-specific bounding box for each proposal.
The architecture is trained in an end-to-end fashion using a multi-task loss.
For training, we used momentum \cite{Qian1999} with a value of 0.9, and a learning rate of
0.0003.
Inception-Resnet V2 \cite{DBLP:journals/corr/SzegedyVISW15} pretrained on ImageNet \cite{imagenet_cvpr09} was used as the feature extractor.

\vspace{2mm}\noindent\textbf{R-FCN} \cite{dai16rfcn}
is similar to Faster R-CNN.
However, instead of cropping features from the same layer where region proposals are predicted,
crops are taken from the last layer of features prior to prediction.
The main idea behind this approach is to minimize the amount of per-region computation that must be done.
Inference time is faster than Faster R-CNN but with comparable accuracy.
For training we used RmsProp \cite{tieleman2012lecture} with an initial learning rate of 0.0001, a decay step
of 10000, a decay factor of 0.95, momentum and decay values of 0.9, and epsilon of 1.0.
As with Faster R-CNN, we used Inception-Resnet V2 pretrained on ImageNet as the feature extractor.

\vspace{2mm}\noindent\textbf{SSD} \cite{Liu2016} uses a single feed-forward convolutional network to directly predict classes and anchor offsets without requiring a second stage per-proposal classification operation.
The predictions are then followed by a non-maximum suppression step to produce the final detections.
This architecture uses the same training strategies as Faster R-CNN.
We used Resnet101 \cite{kaiming2016CVPR} pretrained on ImageNet as the feature extractor.

For our DR dataset, we generated 100K images containing no more than 14 cars each.
As described in the previous section,
each car instance was randomly picked from a set of 36 models, and a random texture from 8K choices was applied.
For comparison, we used the VKITTI dataset
composed of 2.5K images generated by the Unity 3D game engine \cite{gaidon2016CVPR}.
(Although our approach uses more images, these images come essentially for free since they are generated automatically.)
Note that this VKITTI dataset was specifically rendered with the intention of recreating as closely as possible 
the original real-world KITTI dataset (used for testing).

During training, we applied the following data augmentations:
random brightness,
random contrast, and
random Gaussian noise.
We also included more classic augmentations to our training process, such as
random flips,
random resizing,
box jitter,
and random crop.
For all architectures, training was stopped when performance on the test set saturated to avoid overfitting, and only the
best results are reported.
Every architecture was trained on a batch size of~4 on an NVIDIA DGX Station.  (We have also trained on a Titan X with a smaller batch size with similar results.)

For testing, we used 500 images taken at random
from the KITTI dataset \cite{Geiger2012CVPR}.
Detection performance was evaluated using average precision (AP) \cite{Everingham15},
with detections judged to be true/false positives by measuring
bounding box overlap with intersection over union (IoU) at least 0.5.
In these experiments we only consider objects for evaluation that have
a bounding box with height greater than 40 pixels and truncation lower than 0.15, as in \cite{Geiger2012CVPR}.\footnote{These restrictions are the same as the ``easy difficulty'' on the KITTI Object Detection Evaluation website, \url{http://www.cvlibs.net/datasets/kitti/eval_object.php}.}

Table~\ref{tbl:comparison_dectectors} compares the performance of the three architectures when trained on VKITTI versus our DR dataset.
The highest scoring method, Faster R-CNN, performs better with VKITTI than with our data.
In contrast, the other methods achieve higher AP with our DR dataset than with VKITTI, despite the fact that the latter is closely correlated with the test set, whereas the former are randomly generated.

\begin{table}
    \scriptsize
    \centering
    \begin{tabular}{p{2.5cm}>{\centering\arraybackslash}p{1.2cm}>{\centering\arraybackslash}p{1cm}}
        \toprule
        \scriptsize
         Architecture &  \textit{VKITTI} \cite{gaidon2016CVPR} & \textit{DR} (ours) \\
        \midrule
        Faster R-CNN~\cite{faster-rcnn2015NIPS}
        					   & \textbf{79.7} & 
                     78.1 %
                     \\
		R-FCN~\cite{dai16rfcn} & 64.6 & \textbf{71.5}\\
		SSD~\cite{Liu2016}     & 36.1 & \textbf{46.3}\\
        \bottomrule
        \\
    \end{tabular}
    \caption{Comparison of three different state-of-the-art object detectors
    trained on Virtual KITTI versus our DR-generated dataset.  Shown are AP@0.5 numbers from a subset of the real-world KITTI dataset
    \cite{Geiger2012CVPR}.}
    \label{tbl:comparison_dectectors}
\end{table}

\begin{figure*}
    \centering
    \begin{tabular}{ccc}
      ground truth & VKITTI & DR (ours) \\
      \includegraphics[width=0.3\textwidth]{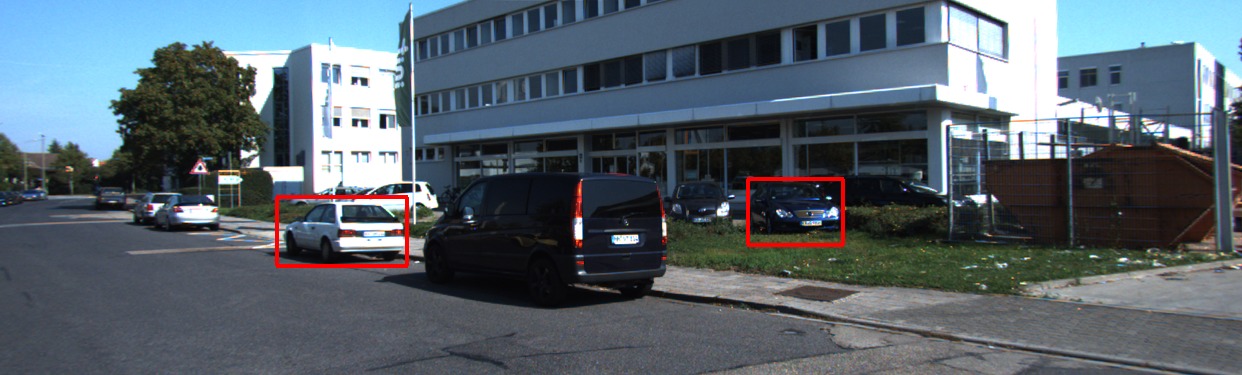} &
      \includegraphics[width=0.3\textwidth]{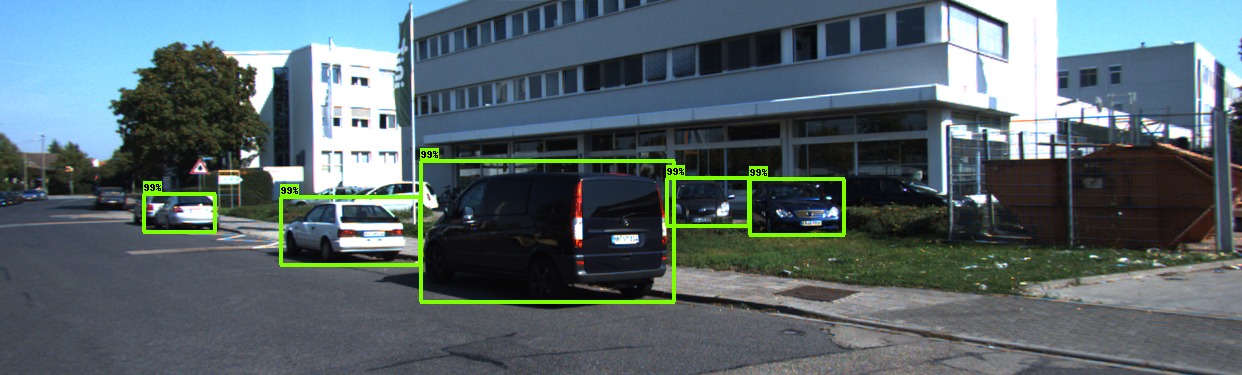} &
      \includegraphics[width=0.3\textwidth]{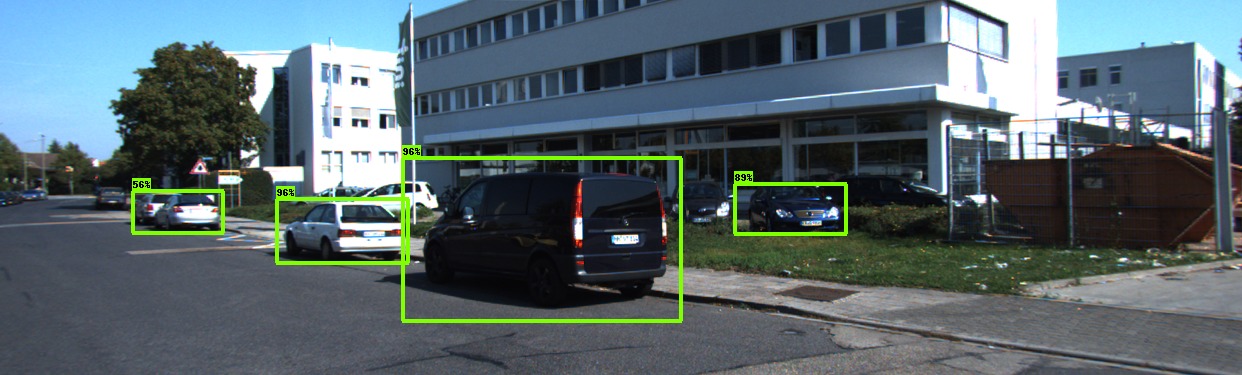} \\
      \includegraphics[width=0.3\textwidth]{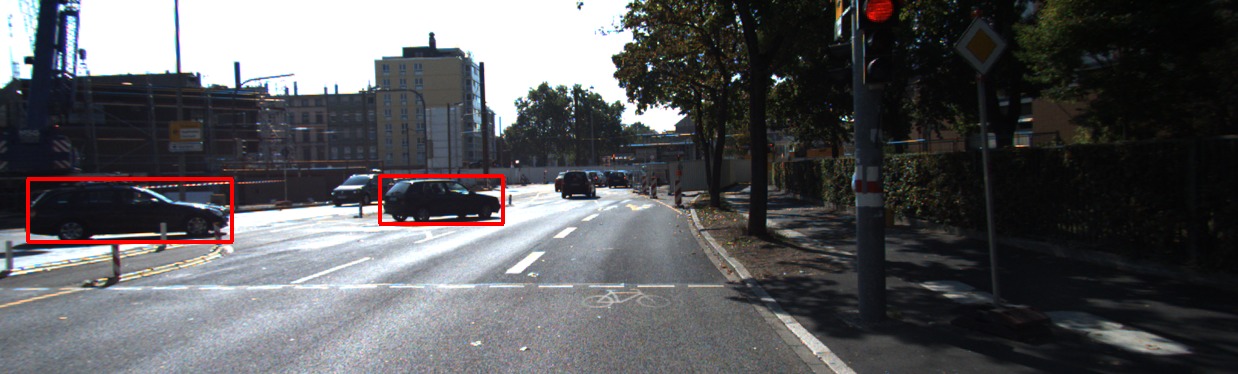} &
      \includegraphics[width=0.3\textwidth]{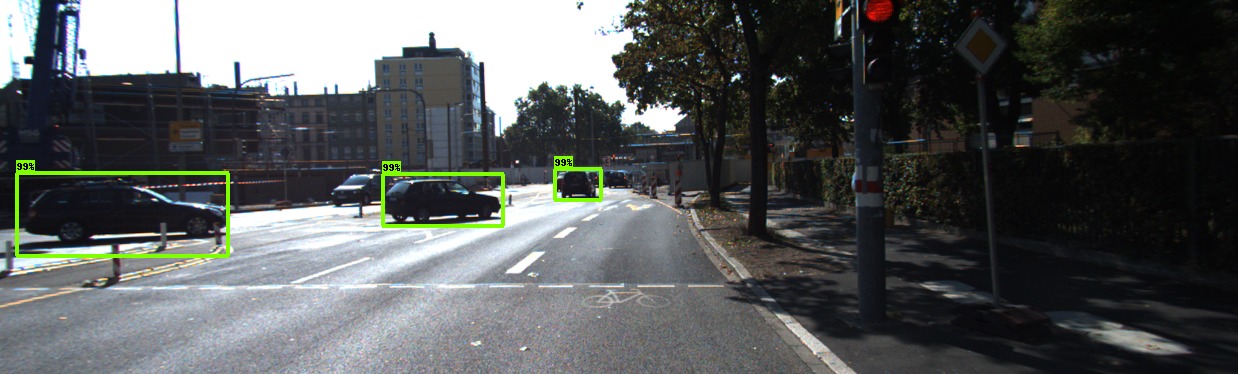} &
      \includegraphics[width=0.3\textwidth]{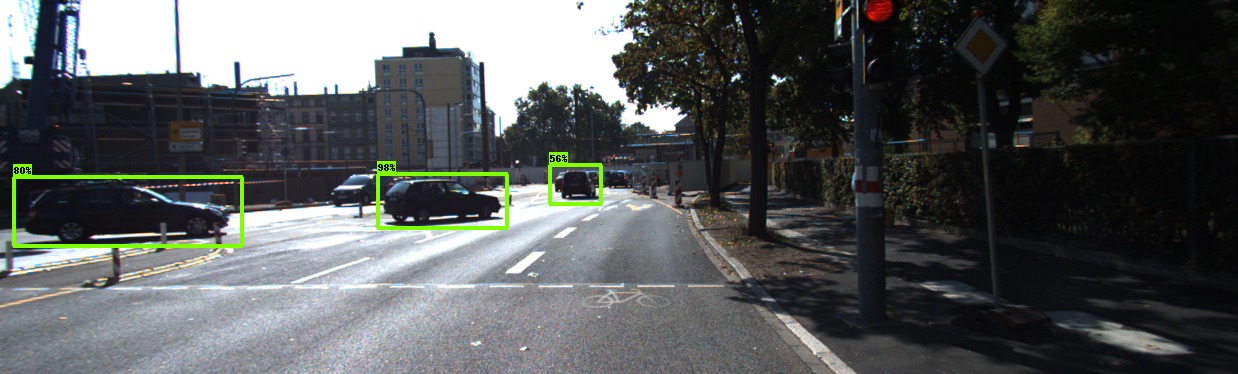} \\
      \includegraphics[width=0.3\textwidth]{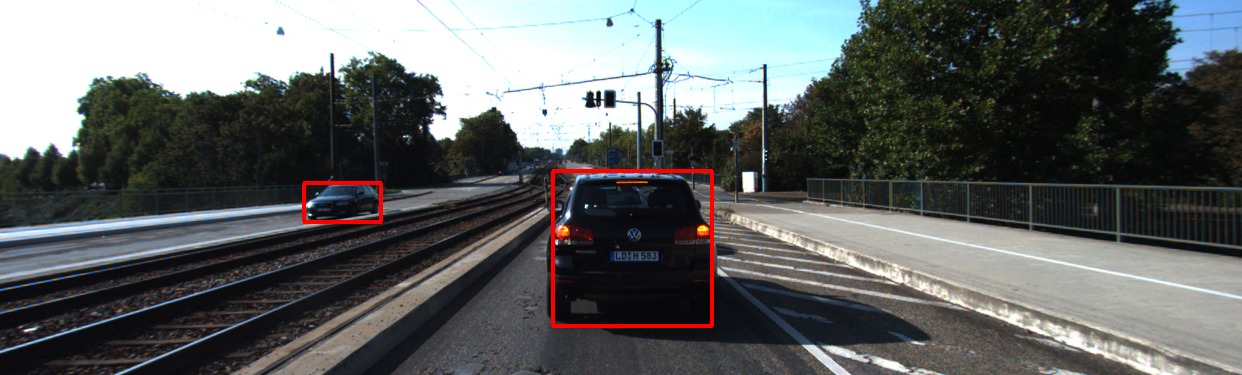} &
      \includegraphics[width=0.3\textwidth]{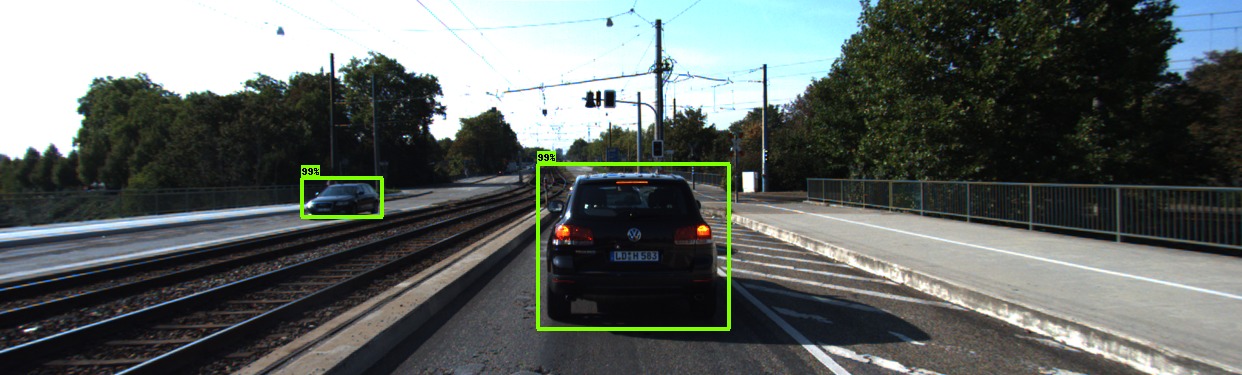} &
      \includegraphics[width=0.3\textwidth]{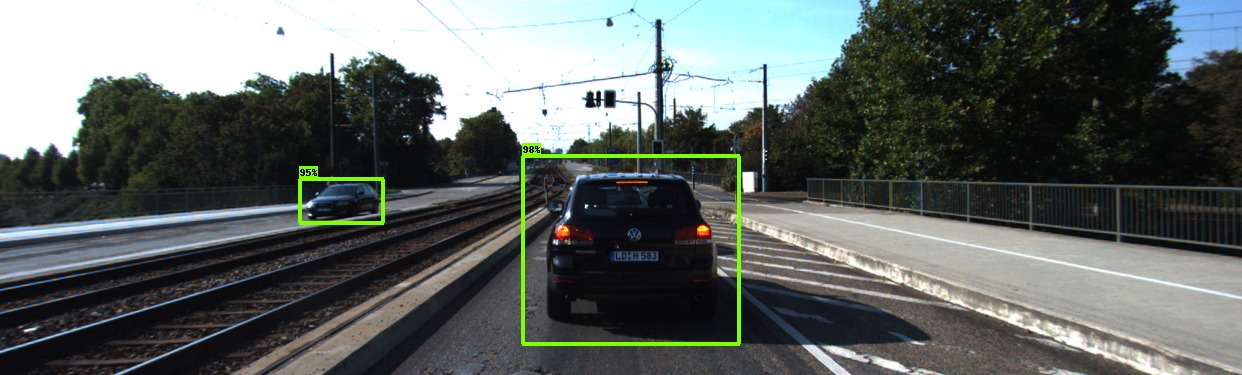} \\
      \includegraphics[width=0.3\textwidth]{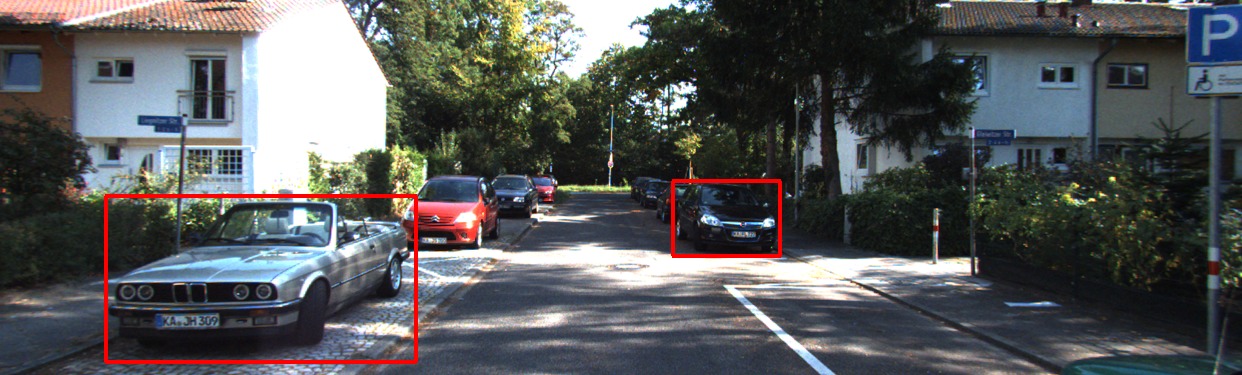} &
      \includegraphics[width=0.3\textwidth]{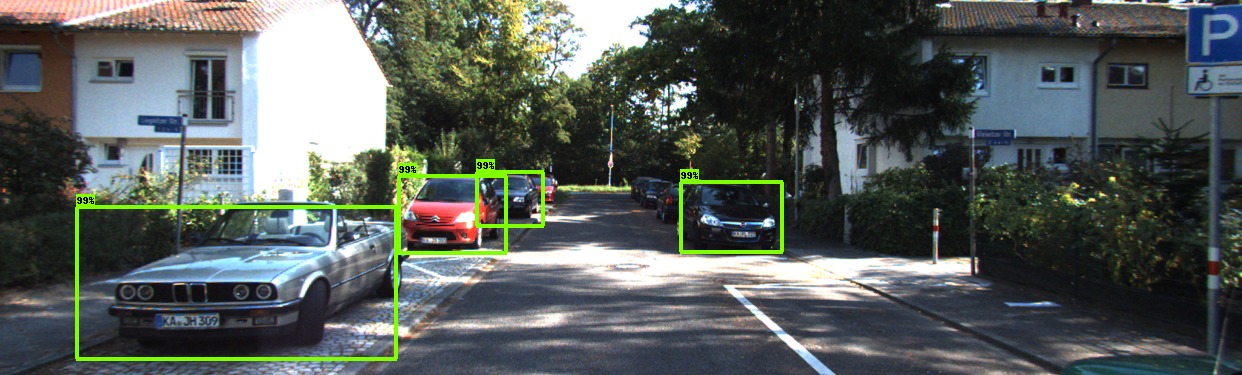} &
      \includegraphics[width=0.3\textwidth]{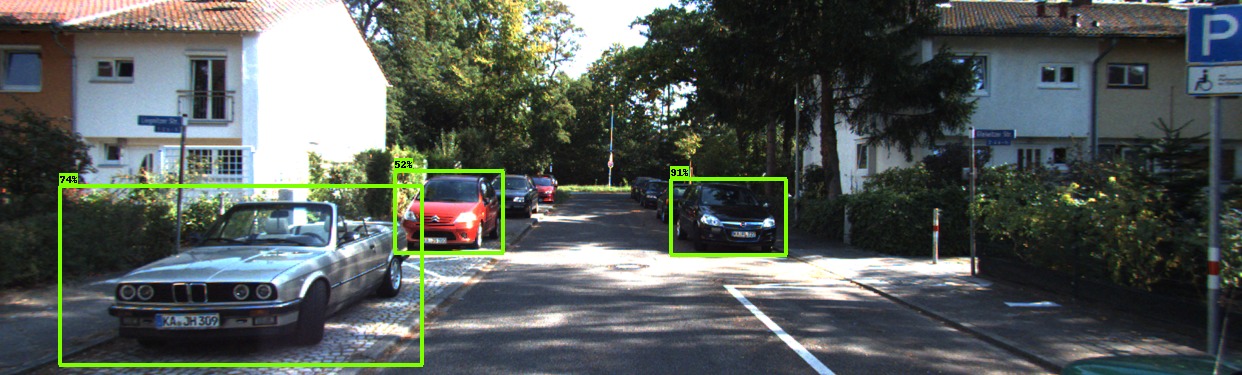} \\

    \end{tabular}
    \caption{Bounding box car detection on real KITTI images using Faster-RCNN trained only on synthetic data, either on the VKITTI dataset (middle) or our DR dataset (right).  Note that our approach achieves results comparable to VKITTI, despite the fact that our training images do not resemble the test images.}
    \label{fig:car_detection}
\end{figure*}

Fig.~\ref{fig:car_detection} shows sample results from the best detector 
(Faster R-CNN) on the KITTI test set, after being trained on either our DR or VKITTI datasets.  
Notice that even though our network has never seen a real image 
(beyond pretraining of the early layers on 
ImageNet),
it is able to successfully detect most of the cars.
This surprising result illustrates the power of such a simple technique as DR for bridging the reality gap.

\begin{figure}
    \begin{center}
        \includegraphics[width=1\linewidth]{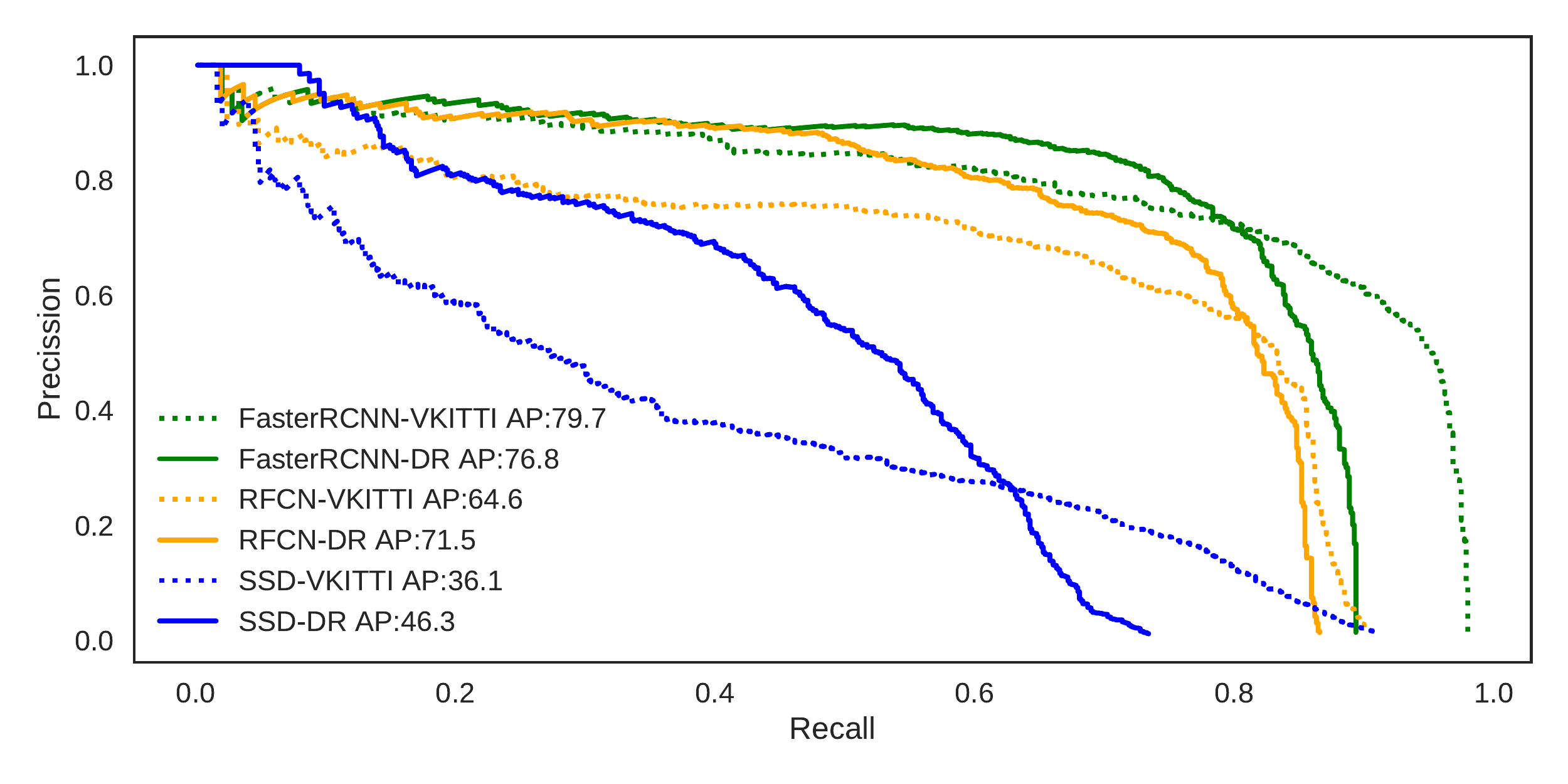}
    \end{center}
   \caption{Precision-recall curves for all trained models on DR (solid)
            and VKITTI (dashed).}
    \label{fig:pr_imagenet}
\end{figure}

Precision-recall curves for
the three architectures trained on both DR and VKITTI are shown in
Fig.~\ref{fig:pr_imagenet}.
From these plots observe that DR actually consistently achieves higher precision than VKITTI for most values of recall for all architectures.  
This helps to explain the apparent inconsistency in Table~\ref{tbl:comparison_dectectors}.  

On the other hand, for high values of recall, DR consistently achieves lower precision than VKITTI.  
This is likely due to a mismatch between the distribution of our DR-based data and the real KITTI data.
We hypothesize that our simplified DR procedure prevents some variations observed in the test set from being generated.  
For example, image context is ignored by our procedure, so that the structure inherent in parked cars is not taken into account. 

\begin{figure}
	\begin{center}
   		\includegraphics[width=1\linewidth]{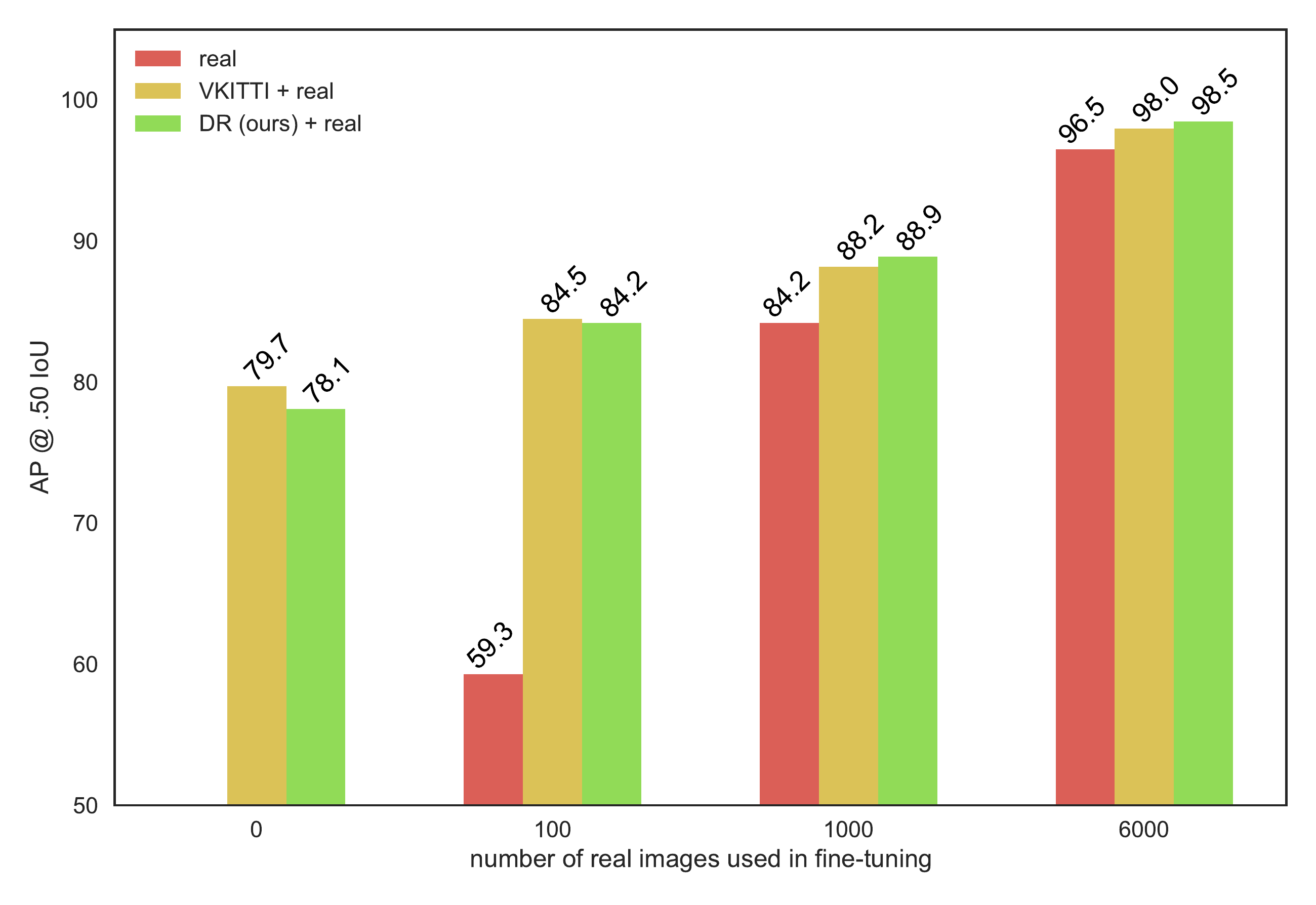}
	\end{center}
   \caption{Results (AP@0.5) on real-world KITTI of Faster R-CNN fine-tuned on real images after training on synthetic images (VKITTI or DR), as a function of the number of real images used.  For comparison, results after training only on real images are shown. 
   }
	\label{fig:datasetcompare}
\end{figure}

In an additional experiment, we explored the benefits of fine-tuning \cite{YosinskiCBL14} on real images after first training on synthetic images.  
For fine-tuning, the learning rate was decreased by a factor of ten while keeping the rest of the hyperparameters unchanged,
the gradient was allowed to fully flow from end-to-end,
and
the Faster R-CNN network was trained until convergence.
Results of VKITTI versus DR as a function of the amount of real data used is shown in Fig.~\ref{fig:datasetcompare}.  
(For comparison, the figure also shows results after training only on real images at the original learning rate of 0.0003.)
Note that DR surpasses VKITTI as the number of real images increases, likely due to the fact that the advantage of the latter becomes less important as real images that resemble the synthetic images are added.  With fine-tuning on all 6000 real images, our DR-based approach achieves an AP score of 98.5, which is better than VKITTI by 1.6\% and better than using only real data by 2.1\%.  

\subsection{Ablation study}

To study the effects of the individual DR parameters, this section presents an ablation study by systematically omitting them one at a time.  For this study, we used Faster R-CNN \cite{faster-rcnn2015NIPS} with Resnet V1 \cite{kaiming2016CVPR}
pretrained on ImageNet as a feature extractor \cite{imagenet_cvpr09}.
For training we used 50K images, momentum \cite{Qian1999} with a value of 0.9, and a learning rate of
0.0003.
We used the same performance criterion as in the earlier detection evaluation,
namely, AP@0.5 on the same KITTI test set.

\begin{figure}
	\begin{center}
        \includegraphics[width=1\linewidth]{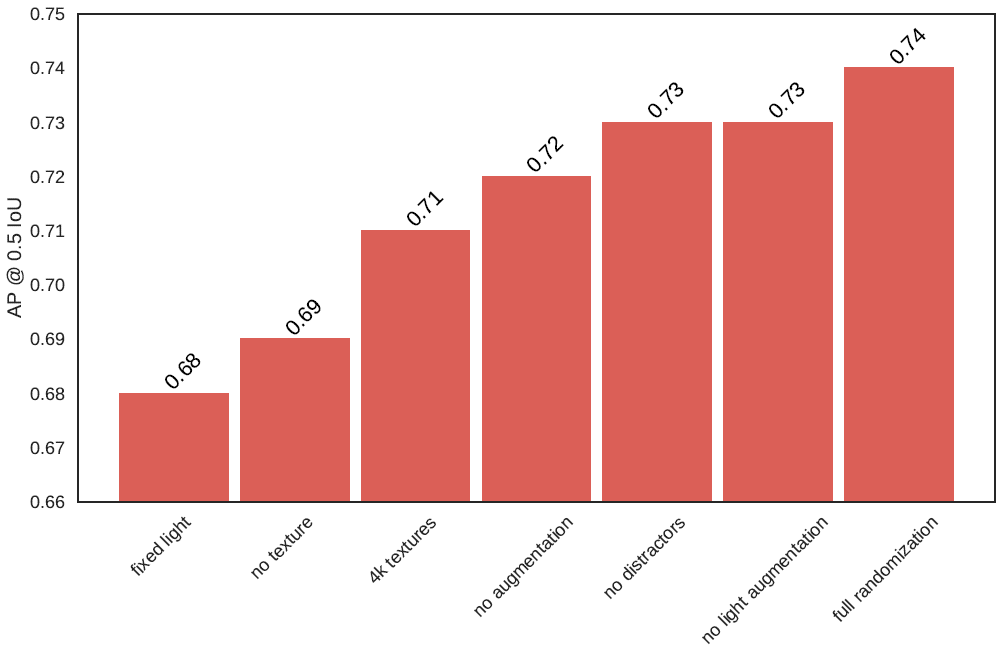}
	\end{center}
   \caption{
   Impact upon AP by omitting individual randomized components from the DR data generation procedure.
  }
	\label{fig:ablation}
\end{figure}

Fig.~\ref{fig:ablation} shows the results of omitting individual random components of the DR approach, showing the effect of these on detection performance, compared with the baseline (`full randomization'), which achieved an AP of 73.7.
These components are described in detail below.

\vspace{2mm}\noindent\textbf{Lights variation.}
When the lights were randomized but the brightness and contrast were turned off (`no light augmentation'), the AP dropped barely, to 73.6. 
However, the AP dropped to 67.6 when the detector was trained on
a fixed light (`fixed light'), thus showing the importance of using random lights.

\vspace{2mm}\noindent\textbf{Texture.}
When no random textures were applied to the object of interest (`no texture'), the AP fell to 69.0.
When half of the available textures were used (`4K textures'), that is, 4K rather than 8K,
the AP fell to 71.5.

\vspace{2mm}\noindent\textbf{Data augmentation.}
This includes randomized contrast, brightness, crop, flip,
resizing, and additive Gaussian noise.
The network achieved an AP of 72.0 when augmentation was turned off (`no augmentation').

\vspace{2mm}\noindent\textbf{Flying distractors.} 
These force the network to learn to ignore nearby patterns and to deal with partial occlusion of the object of interest.
Without them (`no distractors'), the performance decreased by 1.1\%.

\subsection{Training strategies}

We also studied the importance of the pretrained weights, 
the strategy of freezing these weights during training, and the impact of dataset size upon performance.

\vspace{2mm}\noindent\textbf{Pretraining.}\hspace{1mm} For this experiment we used Faster R-CNN with Inception ResNet V2 as the feature extractor.
To compare with pretraining on ImageNet (described earlier), this time we initialized the network weights using COCO~\cite{Lin2014COCO}.
Since the COCO weights were obtained by training on a dataset that already contains cars, we expected them to be able to perform with some reasonable amount of accuracy.  We then trained the network, starting from this initialization, on both the VKITTI and our DR datasets.

The results are shown in Table~\ref{tab:cococo}.  First, note that the COCO weights alone achieve an AP of only 56.1, showing that the real COCO images are not very useful for training a network to operate on the KITTI dataset.  This is, in fact, a significant problem with networks today, namely, that they often fail to transfer from one dataset to another.  Synthetic data, and in particular DR, has the potential to overcome this problem by enabling the network to learn features that are invariant to the specific dataset.  Our DR approach achieves an AP of 83.7 when used to train the network pretrained on COCO, compared with an AP of 79.7 achieved by training on VKITTI.  Thus, DR improves upon the performance of COCO and COCO+VKITTI by 49\% and 5\%, respectively.

\begin{table}%
    \scriptsize
    \centering
	\begin{tabular}{ccc}
			\toprule
			\scriptsize
		COCO & COCO+VKITTI & COCO+DR (Ours) \\
			\midrule
      56.1 & 79.7 & \textbf{83.7} \\
			\bottomrule
			\\
\end{tabular}
\caption{AP performance on KITTI using COCO initialized weights.  Shown are the results from no additional training, training on VKITTI, and training on DR.  Note that our DR approach yields the highest performance.}
\label{tab:cococo}
\end{table}

\vspace{2mm}\noindent\textbf{Freezing weights.}\hspace{1mm} Hinterstoisser {\em et al.}
\cite{hinterstoisser2017arx:pretrain} have recently proposed to freeze the weights of the early network layers (\emph{i.e.}, the feature extraction weights pretrained on ImageNet) when learning from synthetic data.  
To test this idea, we trained Faster R-CNN and R-FCN using the same hyperparameters as in Section~\ref{sec:detection}, except
that we froze the weights initialized from ImageNet rather than allowing them to update during training.  
As shown in Table~\ref{tbl:freezing}, we found that freezing the weights in this manner actually decreased rather than increased performance, contrary to the results of \cite{hinterstoisser2017arx:pretrain}.  This effect was significant, degrading results by as much as 13.5\%.
We suspect that the large variety of our data enables full training to adapt these weights in an advantageous manner, and therefore freezing the weights hinders performance by preventing this adaptation.

\begin{table}
    \scriptsize
    \centering
    \begin{tabular}{lcc}
        \toprule
        \scriptsize
         architecture &  freezing \cite{hinterstoisser2017arx:pretrain} & full (Ours)	 \\
        \midrule
        \textit{Faster R-CNN} \cite{faster-rcnn2015NIPS}
        					& 66.4 & \textbf{78.1}\\
		\textit{R-FCN} \cite{dai16rfcn}
							& 69.4 & \textbf{71.5}\\
        \bottomrule
        \\
    \end{tabular}
    \caption{Comparing freezing early layers {\em vs.}\ full learning for different detection
    architectures.}
    \label{tbl:freezing}
\end{table}

\begin{figure}
  \begin{center}
        \includegraphics[width=1\linewidth]{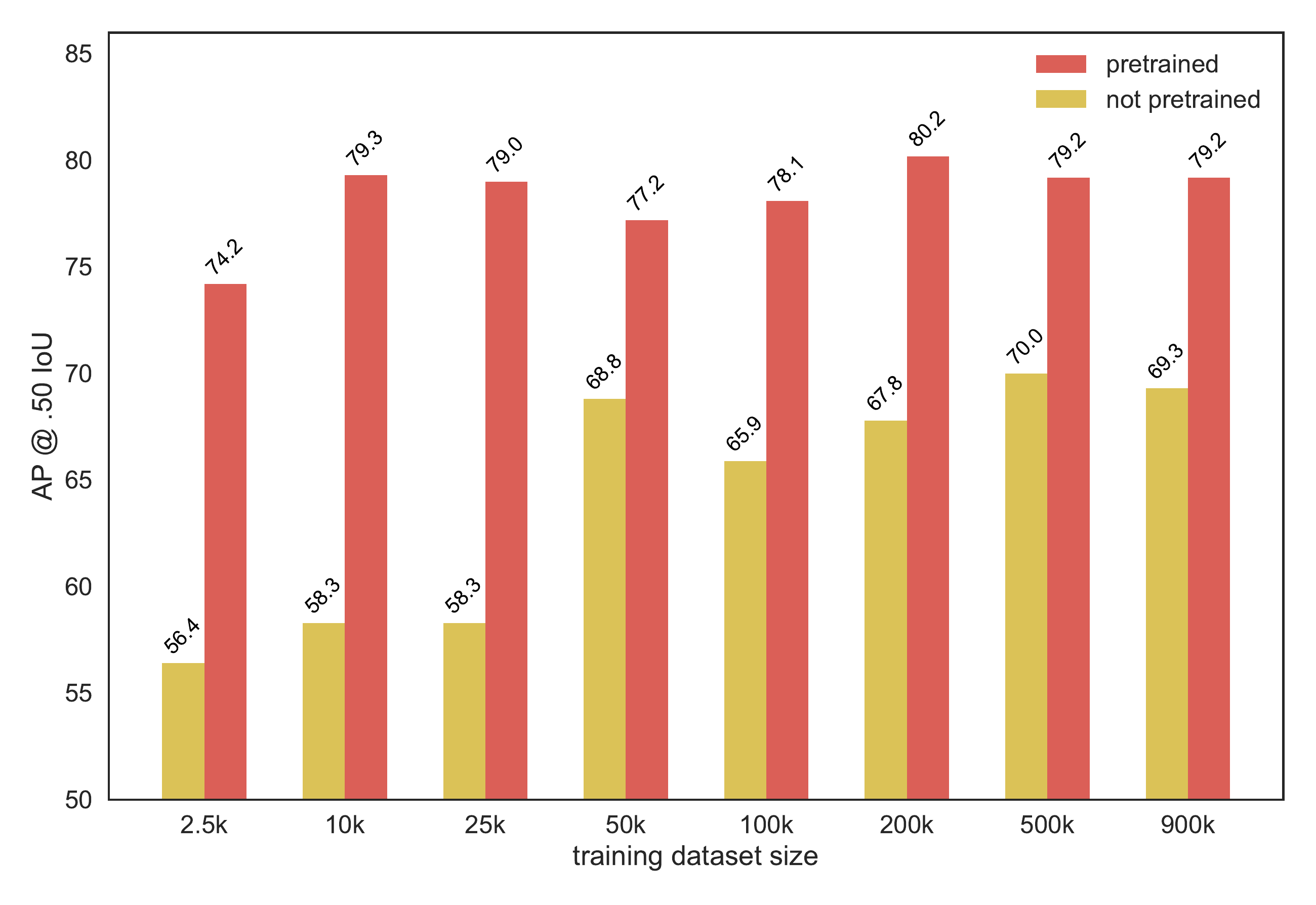}

  \end{center}
   \caption{Performance of Faster R-CNN as a function of the number of training images used, for both pretrained weights using ImageNet (red) and randomly initialized weights (yellow).  
   }
  \label{fig:data_size}
\end{figure}

\vspace{2mm}\noindent\textbf{Dataset size.}\hspace{1mm} 
For this experiment we used the Faster R-CNN architecture with either randomly initialized weights or the Inception ResNet V2 weights. 
Using these two models, we explored the influence of dataset size upon
prediction performance. 
We used the same data generation procedure explained earlier.
The results, shown in Fig.~\ref{fig:data_size}, surprisingly reveal that performance saturates after only about 10K of training images with pretrained weights or after about 50K without pretraining.
Conflicting somewhat with the findings of \cite{McCormac:imagenet_init_beat}, we discovered that pretraining helps significantly even up to 1M images.  
This result can be explained by the fact that our training images are not photorealistic.

\section{Conclusion}
\label{sec:conclusion}

We have demonstrated that domain randomization (DR) is an effective technique to bridge the reality gap.
Using synthetic DR data alone, we have trained a neural network to accomplish complex tasks like object detection with performance comparable to more labor-intensive (and therefore more expensive) datasets.
By randomly perturbing the synthetic images during training, DR intentionally abandons photorealism to force the network to learn to focus on the relevant features.
With fine-tuning on real images, we have shown that DR both outperforms more photorealistic datasets and improves upon results obtained using real data alone.
Thus, using DR for training deep neural networks is a promising approach to leveraging the power of synthetic data.
Future directions that should be explored include using more object models, adding scene structure (\emph{e.g.}, parked cars), applying the technique to heavily textured objects (\emph{e.g.}, road signs), and further investigating the mixing of synthetic and real data to leverage the benefits of both.  

%
%
%
%
%
%
%
%
%
%
%
%
%
%
%

\section*{ACKNOWLEDGMENT}

 We would like to thank Jan Kautz, Gavriel State, Kelly Guo, Omer Shapira, Sean Taylor, Hai Loc Lu, Bryan Dudash, and Willy Lau for the valuable insight they provided to this work.

{\small
\bibliographystyle{ieee}
\bibliography{cvpr2018-cars}
}

\end{document}